\documentclass{article}

\usepackage[preprint]{neurips_2019}

\usepackage[utf8]{inputenc} 
\usepackage[T1]{fontenc}    
\usepackage{hyperref}       
\usepackage{url}            
\usepackage{booktabs}       
\usepackage{amsfonts}       
\usepackage{nicefrac}       
\usepackage{microtype}      
\usepackage{color}

\usepackage{algorithm,algorithmic}
\usepackage{mathtools}
\usepackage{multirow}
\usepackage{multicol}
\usepackage{float}

\usepackage{sidecap}
\sidecaptionvpos{figure}{c}
\usepackage{graphicx,wrapfig,lipsum}

\title{Overcoming Catastrophic Forgetting by Neuron-level Plasticity Control}

\author{%
Inyoung Paik, Sangjun Oh, Tae-Yeong Kwak\\
Deep Bio Inc., Seoul, Republic of Korea \\
\textit{(iypaik, juneoh, tykwak)@deepbio.co.kr} \\
\AND
Injung Kim \\
Handong Global University, Pohang, Republic of Korea \\
\textit{ijkim@handong.edu} \\
}

\begin{document}

\maketitle

\begin{abstract}
To address the issue of catastrophic forgetting in neural networks, we propose a novel, simple, and effective solution called \textit{neuron-level plasticity control (NPC)}. While learning a new task, the proposed method preserves the knowledge for the previous tasks by controlling the plasticity of the network at the neuron level. NPC estimates the importance value of each neuron and consolidates important \textit{neurons} by applying lower learning rates, rather than restricting individual \textit{connection weights} to stay close to certain values. The experimental results on the incremental MNIST (iMNIST) and incremental CIFAR100 (iCIFAR100) datasets show that neuron-level consolidation is substantially more effective compared to the connection-level consolidation approaches.
\end{abstract}

\section{Introduction}

In the path to realizing artificial general intelligence with deep neural networks, 
catastrophic forgetting remains one of the most fundamental challenges. Gradient descent, the most popular learning algorithm, causes the problem when it is applied to train a neural network for multiple tasks in a sequential manner. When gradient descent optimizes the neural network for the task at hand, the knowledge for the previous tasks is 
\textit{catastrophically} overwritten by the new knowledge.

Since the initial discovery of the problem\citep{classic1}, various approaches have been 
proposed to alleviate catastrophic forgetting in artificial neural networks. One of these 
approaches is to include the data for multiple tasks in every mini-batch. Although such a 
method can be effective in retaining the performance of the previous tasks, it causes an 
overhead to keep the training data for the previous tasks. There have been several attempts to achieve a similar effect using only a limited portion of the previous data, \citep{BioIncrementalLeraning, GEM, Baeysian} or none at all.\citep{LwF, DGR, DM, SeNA, KeepAndLearn}

Another approach is to isolate the parts of the neural network containing previous knowledge, and learn the new task using the other parts of the network. This includes designing dynamic architectures for neural networks,\citep{PathNet, ExpertGate, ExpendableNetwork, HAT, XdG} where the capacity to learn the new task is obtained by assigning different parts of the network to the new task. Note that these methods isolate a portion of \textit{neurons}, rather than \textit{parameters}, to effectively(sometimes, even perfectly) preserve the existing knowledge of the neural network. Our work is related to this approach because the proposed algorithm also learns multiple tasks using different parts of the network, where the unit of the 'part' is the individual \textit{neuron}. 

Weight consolidation is a remarkable step made in this area. Elastic Weight Consolidation(EWC)\citep{EWC} uses the diagonals of the Fisher information matrix to identify and consolidate the parameters, which correspond to the connection weights in neural networks, that are important for the previous tasks. In this way, the network is capable of learning the new task using less important parameters while preserving previously learned knowledge. In Memory Aware Synapses(MAS)\citep{MAS}, the importance value of each weight is measured by the sample mean of the absolute value of the gradient. In Synaptic Intelligence(SI)\citep{SI}, the importance value of each weight is computed by integrating the contribution to the change in loss. Meanwhile, Selfless Sequential Learning(SSL)\citep{SSL} suggests imposing sparsity on neuron activation while learning the preceding tasks to avoid exhausting all capacities. Their work is similar to ours in that both methods focus on neuron-level information. However, in contrast to that the purpose of SSL is to save network capacity for the subsequent tasks, we aim to preserve the knowledge for the previous tasks while learning a new task. Weight consolidation algorithms have drawn much attention, and therefore, have been adopted in many studies\citep{IMM, Rotate}. Recent works\citep{KeepAndLearn, ExpendableNetwork} show that this approach may be used in combination with other methods as a means of regularization.

In this study, we describe the limitation of weight consolidation (or, for purposes of comparison, 'connection-level consolidation') in deep neural networks, and propose an novel algorithm, called neuron-level plasticity control (NPC). As the name suggests, NPC retains existing knowledge by controlling the plasticity of each \textit{neuron} or each filter in a convolutional neural network (CNN). As a result, it is significantly more effective in preserving the knowledge of deep neural networks. Moreover, NPC contains a memory-efficient consolidation algorithm. Most of existing consolidation algorithms restrict individual connection weights to stay close to certain values. optimized for the previous tasks. Such algorithms need to save the weight and importance value of each connection for every task, and therefore, require memory and computation linear to the number of tasks. On the other hand, NPC controls the plasticity of the neurons by simply adjusting their learning rates according to their importance value. The proposed consolidation algorithm removes the overhead to keep the task-specific information. As NPC needs to store only a single importance value per neuron instead of multiple sets of per-task parameter values, the memory requirement remains consistent regardless of the number of tasks.



\section{Neuron-level Versus Connection-level Consolidation}

While the connection-level consolidation algorithms\citep{EWC, IMM, Rotate, SI, MAS} focus on the idea that knowledge is stored in the parameters, which are the \textit{connection weights} for neural networks, less emphasis is given to the \textit{correlation} among these connections. The connection-level consolidation can be represented as the following loss function:  
\begin{equation}
    Loss = \lambda \sum_{k<n} \sum_i W_i (\theta_i - \theta_{i, k})^2
\end{equation}
Where $\theta_{i}$ and $\theta_{i, k}$ denote the $i$-th parameter and its value at the end of the learning of the $k$-th task, respectively. $W_i$ denotes the importance value of $\theta_{i}$ and $\lambda$ is a hyperparameter.

Note that there is an implicit assumption that the weights in neural networks are roughly 
independent, and a neural network can be linearly approximated by its weights. 
However, the structure of deep neural networks is  inherently hierarchical, and therefore, there is a strong correlation among parameters. Thus, the connection weights of a deep network are tightly correlated, as the value of a parameter can affect the importance value of another.

We argue that the neurons, or CNN filters, are more appropriate than the individual 
connections for the basic unit of knowledge in the consolidation of the artificial 
neural network. Conventional connection-level algorithms do not guarantee the preservation of important knowledge expressed by neurons. Even if the learning 
algorithm consolidates some of the connections to an important neuron, the neuron 
may have remaining free incoming connections, the change of which may severely 
affect the knowledge carried by the neuron.

\begin{SCfigure}
\label{fig:connections}
\centering
\includegraphics[height=7cm]{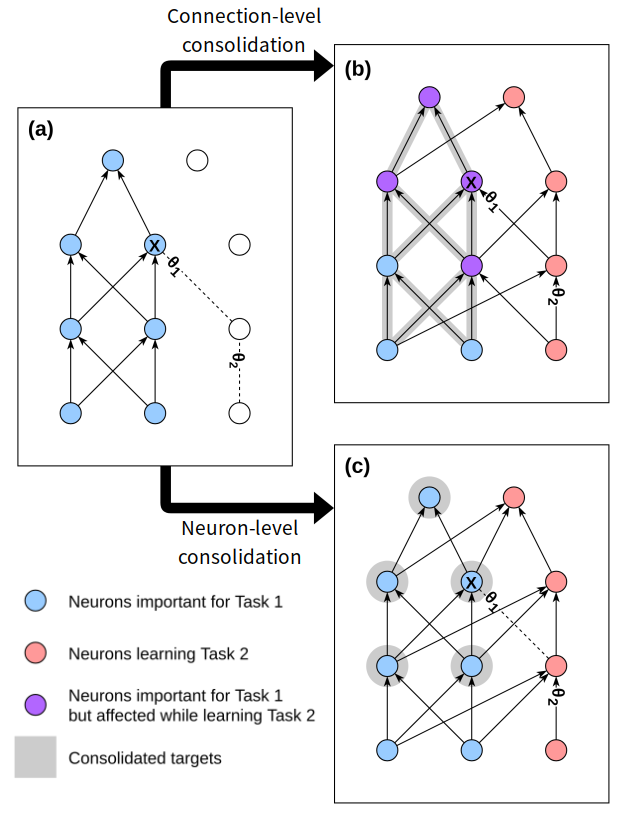}  
\caption{Comparison of connection-level and neuron-level consolidation. (a) The neurons and 
    connections important for Task 1. (b) Connection-level consolidation. The important 
    connections are consolidated, but the neurons can still be affected by other 
    incoming connections that can change while learning Task 2. (c) Neuron-level 
    consolidation. NPC consolidates all incoming connections of the important neurons, 
    which is more effective in preserving the knowledge of the neurons. A similar intuition was pointed out by\citep{SSL, ExpendableNetwork, HAT}}
\end{SCfigure}

Figure \ref{fig:connections} illustrates the limitation of connection-level 
consolidation in deep neural networks more clearly. In the figure, although the neuron $X$ is important for Task 1, changing the value of $\theta_1$ or $\theta_2$ 
individually may not have a significant effect on the output of Task 1, if the value of both $\theta_1$ and $\theta_2$ are close to zero. In such 
a circumstance, due to their low importance values, connection-level algorithms 
would consolidate neither of the two connection parameters. Nonetheless, the neuron X can be seriously affected during the subsequent learning, when both parameters are rapidly increased because of their correlatedness. This problem can be particularly 
serious in convolution layers in which the same filters are shared among multiple positions. Thus, even if the concept of connection-level consolidation can be implemented perfectly, catastrophic forgetting cannot be eliminated completely.

To overcome this problem, we propose controlling the plasticity at the neuron 
level rather than at the connection level as shown in Figure 
\ref{fig:connections}(c). The proposed algorithm, NPC, collectively consolidates 
all the incoming connections of the important \textit{neurons}, including the 
connections that might not individually be evaluated as important. As a result, 
NPC protects the important neurons more effectively from the change of unimportant neurons than the connection-level consolidation algorithms. Note that the 
connection from an unimportant neuron to an important neuron is likely 
to be small, because otherwise, the evaluation algorithm would not have determined the source neuron to be unimportant. In the example shown in Figure 
\ref{fig:connections}, NPC consolidates all the input connections of $X$, and 
as a result, the value of $\theta_1$ remains small, preventing the change of 
$\theta_2$ from affecting $X$ severely. On the other hand, NPC does not 
consolidate a connection whose destination neuron is unimportant, even if the source neuron is important. Therefore, the total number of consolidated connections in the whole network can remain acceptable.
\section{Neuron-level Plasticity Control}

\subsection{Importance Value Evaluation}

To evaluate the importance value of each neuron, we adopt a criterion based on 
Taylor expansion that has been used in the field of network pruning 
\citep{Pruning}. Although there are other methods in network pruning that 
claim to have better performance\citep{Pruning1, Pruning2, Pruning3}, we 
chose the Taylor criterion due to its computational efficiency. The 
Taylor criterion is computed from the gradient of the loss function with 
respect to the neurons, which is computed during back-propagation. 
Therefore, it can easily be integrated into the training process with 
minimal additional computation.

We measure the importance of each neuron by the normalized Taylor criterion as shown in eq. (2) and (3). Then, we take their moving averages as eq. (4) to reduce the fluctuation of the measurements and to improve the learning stability.

\begin{equation}
    c_i^{(t)} = \underset{batch}{average}\, |n^{(t)}_i \frac{dL^{(t)}}{dn^{(t)}_i}|
\end{equation}

\begin{equation}
    \bar{c_i}^{(t)} = \frac{c_i^{(t)}}{\sum_{layer} c_j^{(t)} /N_{layer}}
\end{equation}

\begin{equation}
    C_i^{(0)}=0, \,\, C_i^{(t)} = \delta C_i^{(t-1)}  + (1-\delta) \,\bar{c_i}^{(t)}
\end{equation}

$C_i^{(t)}$ is the importance value of the $i$-th neuron at training step $t$, $n_i$ denotes the activation of the $i$-th neuron. $L$ is the loss, and $N_{layer}$ is the number of nodes on the layer, and $\delta$ is a hyperparameter. If a node is shared in multiple positions, e.g. the convolution filter in CNN, we average the importance values from all positions before computing its absolute value, following the original paper\citep{Pruning}. However, we use the arithmetic mean as in eq. (3), instead of the L2-norm, in order to enforce stricter balance among the layers composed of different number of neurons.

\subsection{Plasticity Control}

The stability-plasticity dilemma is a well-known constraint in both artificial and biological neural systems\citep{Dilemma}. Catastrophic forgetting can be seen as a consequence of the same trade-off problem: attempting to determine the optimal point that maximizes the performance of the neural network for multiple tasks. We control the plasticity of each neuron by applying different learning rates $\eta_i$ for each neuron. If $\eta_i$ is high, the neuron actively learns the new knowledge at the cost of rapidly losing existing knowledge. On the other hand, if $\eta_i$ is low, existing knowledge can be preserved better; however, the neuron will be reluctant to learn new knowledge.

In order to encourage the neural network to find a good stability-plasticity balance, we define two losses as functions of $\eta_i$ that play opposite roles; subsequently, we combine them. The first is the stability-wise loss to minimize the forgetting of existing knowledge. It should be a monotonically increasing function of $\eta_i$ starting at $\eta_i = 0$ and bounded above by the amount of current knowledge. We heuristically approximate the upper bound of the current knowledge by using $a_1 C_i$, where $a_1$ is a scaling constant. In order to make a monotonically increasing function of $\eta_i$, we combine $\mathrm{tanh}(b_1\eta)$ with the upper bound, where $b_1$ is another constant to control the slope of the $\mathrm{tanh}$ function. Consequently, the stability-loss is defined as $a_1 C_i\mathrm{tanh}(b_1\eta_i)$.

The second function is the plasticity-wise loss to decrease the 
reluctance against new knowledge. It is a decreasing function of 
$\eta_i$ that starts from the maximum value at $\eta_i = 0$ and 
decreases monotonically to zero. The upper bound in this case does not regard existing knowledge, and therefore, is unrelated to $C_i$. We thus define the plasticity-wise loss as 
$a_2(1-\mathrm{tanh}(b_2\eta_i))$, where $a_2$ and $b_2$ are 
constants to control the scale and the slope.

To find the balance between stability and plasticity, we choose 
the $\eta_i$ that minimizes the combined loss function eq. (5).

\begin{equation}
\eta_i^* = \underset{\eta_i}{argmin}\,l(\eta_i) =\underset{\eta_i}{argmin}\{ a_1 C_i  \mathrm{tanh}(b_1 \eta_i)+ a_2 (1- \mathrm{tanh}(b_2 \eta_i)) \}
\end{equation}

Setting $dl/d\eta_i = 0$, we get eq. (6), where $\beta = a_2 b_2/a_1 b_1$.

\begin{equation}
    a_1 b_1 C_i sech^2 (b_1 \eta) - a_2 b_2 sech^2 (b_2 \eta) = 0 
\end{equation}
\begin{equation}
    \iff \frac{cosh (b_2 \eta)}{cosh (b_1 \eta)} = \sqrt{\frac{a_2 b_2}{a_1 b_1 C_i}} = \sqrt{\frac{\beta}{C_i}}
\end{equation}

\begin{algorithm}[tb]
\caption{Neuron-level Plasticity Control (NPC)}
\label{NPC}
\begin{algorithmic}
   \STATE $f$ : neural network model
   \STATE $n_i$ : $i$-th neuron in $f$
   \STATE $w_{ji}$ : weight of connection from $n_j$ to $n_i$
   \STATE $\eta_{max}$ : upper bound of learning rate
   \STATE $\alpha, \beta$ : hyperparameters controlling learning rate
   \STATE $\delta$ : decay rate of the importance value of each neuron
   \STATE $C_i$ : importance value of $i$-th neuron
   \STATE
   \STATE $C_i \leftarrow 0, \,\, \forall i$
   \FOR{$input,\, label$ {\bfseries in} full training dataset}
   
   \STATE $y \leftarrow f(input)$
   \STATE $L \leftarrow CrossEntropy(y, label)$
   \FOR{$n_i$ {\bfseries in} $f$}
   \STATE $c_i \leftarrow \underset{batch}{average}\, |n_i \frac{dL}{dn_i}|$
   \STATE $\bar{c_i} \leftarrow \frac{c_i^{(t)}}{\sum_{layer} c_j^{(t)} /N_{layer}}$
   \STATE $C_i \leftarrow \delta C_i  + (1-\delta) \,\bar{c_i}$
   \STATE $\eta_i \leftarrow min(\eta_{max}, \alpha \sqrt{max(\sqrt{\frac{\beta}{C_i}}-1, 0)}) $
   \STATE $w_{ji} \leftarrow w_{ji} - \eta_i \frac{dL}{dw_i},\, \forall j$ 
   \ENDFOR
   \ENDFOR
\end{algorithmic}
\end{algorithm}

The nature of function $cosh (b_2 \eta)/cosh (b_1 \eta)$ depends 
heavily on whether $b_1 \geq b_2$ or $b_1 < b_2$. We set $b_1 < b_2$ as a constraint, since otherwise $\eta_i$ becomes a simple step function.

Let us first assume that $C_i \geq \beta$. we apply 
the Taylor approximation to solve eq. (7) because there is no 
closed-form inverse function of $cosh (b_2 x)/cosh (b_1 x)$. 
Given that $cosh$ is an even function, only the even degree terms 
remain, as shown in eq. (8).
\begin{equation}
    \frac{cosh (b_2 \eta_i)}{cosh (b_1 \eta_i)} = 1 + (b_2^2 - b_1^2) \eta_i^2 + O(\eta_i^4) = \sqrt{\frac{\beta}{C_i}}
\end{equation}
Assuming $O(\eta_i^4) \approx 0$, the 
solution of eq. (8) is the same as eq. (9), where 
$\alpha = 1/\sqrt{b_2^2-b_1^2}$.
\begin{equation}
    \eta_i^* = \sqrt{\frac{\sqrt{\frac{\beta}{C_i}}-1}{b_2^2-b_1^2}} = \alpha \,\sqrt{\sqrt{\frac{\beta}{C_i}}-1}
\end{equation}

In case of $C_i > \beta$, $l(\eta_i)$ is strictly increasing w.r.t. 
$\eta_i$ , which leads to $\eta_i^* = 0$. Note that $\eta_i^*=0$ at $C_i = \beta$ in eq. (9), which makes the two functions continuously connected. Combining the two cases where $C_i > \beta$ and $C_i \leq \beta$, respectively, the solution of eq. (6) is given by eq. (10), where $\alpha, \beta >0$ are hyperparameters.

\begin{equation}
    \eta_i^* = \alpha \, \sqrt{max(\sqrt{\frac{\beta}{C_i}}-1, 0)}
\end{equation} 

In eq. (10), a larger $C_i$ draws a smaller $\eta_i^*$, thereby consolidating the important neurons in the subsequent learning. However, if $C_i = 0$, then $\eta_i^*$ diverges. This is explainable from the perspective of the plasticity-stability dilemma: if a neuron has no knowledge at all, it is desirable to learn the new knowledge as much as possible without considering the loss of existing knowledge. However, this can not be applied to reality. Therefore, we set an upper bound of the learning rate $\eta_{max}$. The final solution of eq. (6) is given by eq. (11).

\begin{equation}
    \eta_i^* = min(\eta_{max}, \alpha \, \sqrt{max(\sqrt{\frac{\beta}{C_i}}-1, 0)})
\end{equation}

Algorithm \ref{NPC} summarizes the NPC algorithm. Considering that $C_i$ is also simply computed from the activation and the gradient, which are computed by the back-propagation algorithm, the overhead to implement NPC is minimal.

\subsection{Instance Normalization}

Batch normalization (BN) plays a key role in the training of deep 
neural networks\citep{BatchNorm}. However, the vanilla batch 
normalization does not work well in continual learning environments, 
because the mean and the variance are heavily affected by the 
transition of tasks. There are a few alternatives available in such 
cases, such as conditional batch normalization\citep{ConditionalBN} and 
virtual batch normalization\citep{VirtualBN}. However, they does not fit the objective of the NPC as they maintain task-specific information. Therefore, we apply a instance normalization\citep{InstanceNorm} without  
affine transforms. Given that instance 
normalization can be applied to each sample independently, it operates 
without any special manipulation of model parameters, not only at the 
training time but also at the test time.

\section{Experiments}


We experimented on \textit{incremental} versions of MNIST\citep{MNIST} and CIFAR100\citep{CIFAR} datasets, where the datasets containing $X$ classes were divided into $K$ subsets of $X/K$ classes, each of which is classified by the $k$-th task. $K$ was set to $5$ for MNIST and $10$ for CIFAR100. For preprocessing, we applied random cropping with padding size of 4 for both datasets, and an additional random horizontal flip for the incremental CIFAR100 (iCIFAR100) dataset. For consistency, we redefined the unit of \textit{one epoch} in all experiments as the cycle in which the total number of train data was seen. For example, as the original MNIST dataset has 60,000 training samples, we defined one epoch of the iMNIST dataset as the processing of the 12,000 task-specific samples five times. With this new definition of an epoch, we trained the models for 30 epochs on each subset of iMNIST or iCIFAR100. We used the first five subsets of iCIFAR100 in the experiment. All experiments were performed on a server with 2 NVIDIA Tesla P40 GPUs.

We used a simple CNN with 3 convolutional layers with (64, 256, 128) channels, and 2 fully connected layers with ($128\cdot4\cdot4$, 512) nodes. Each convolutional layer consists of convolution, Instance normalization, ReLU activation, and (2,2) max pooling. Dropout\citep{dropout} of rate 0.2 is applied between two fully connected layers. The cross-entropy loss for each task was computed from only the output nodes belonging to the current task.

For direct comparison between neuron-level and parameter-level consolidation, we implemented an alternative algorithm called 'Connection-level Plasticity Control(CPC)' that is almost the same as NPC except that it consolidates the network at the connection level. Note that producing a neuron-level counterpart of weight consolidation is not trivial, since restricting individual neurons to stay close to certain values is not appropriate.

We set $\alpha = 0.1, \beta = 0.7, \eta_{max}=0.1,$ and used $\delta = 10^{-3}$ in iMNIST and $5 \cdot 10^{-4}$ in iCIFAR100 experiment respectively. We used a plain SGD optimizer with mini-batch size of 512, and learning rate 0.05 where NPC(or CPC) is not used. We compared our methods with EWC\citep{EWC}, SI\citep{SI}, MAS\citep{MAS}, SSL\citep{SSL}, and baseline SGD with fine-tuning. In the hyperparameter search, we started with the value used by the author, and then searched with a unit of 10 from $10^{-3}$ to $10^3$ based on average validation accuracy, and then performed some further searches around it. All experiments are averaged over 3 runs, but Figure (2) is a visualization of single run. Please see Table (1), (2) for further details including average accuracy and standard error. Note that we used different hyperparameters for iMNIST and iCIFAR100 experiments in some algorithms, to obtain best performance for each algorithm.

In both experiments, NPC performed better than the connection-level consolidation algorithms. In particular, we found that CPC is very inefficient at preserving old knowledge, even though it consolidates similar numbers of weights with NPC. This shows that the neurons are more appropriate than the connections as the units of neural network consolidations.

Additionally, we measured the change in the activation of the neurons on the second top layer after learning a subsequent task to see whether NPC successfully consolidates important neurons or not. First, we trained a CNN for Task 1 of iCIFAR100 for 30 epochs and recorded the neuron activation values of the second top neurons(just before the final classifier) extracted from randomly chosen 256 samples. (512 neurons $\times$ 256 samples = 131,072 data points in total.) Then, we trained the CNN for Task 2 for another 30 epochs. Finally, we measured the change in the neuron activation values from the same sample. Figure 3 displays the results. Without consolidation, there was no meaningful correlation between the change in neuron activation and their importance values. However, using NPC, the average change of activation of all neurons was 0.383. The average change of activation of the most important 10\% of the neurons was only 0.094, while that of the least important 10\% of neurons was 0.667. These results suggest that NPC successfully preserved the neurons important for

\begin{figure}[H]
\vskip 0.2in
\begin{center}
\centerline{\includegraphics[width=\columnwidth]{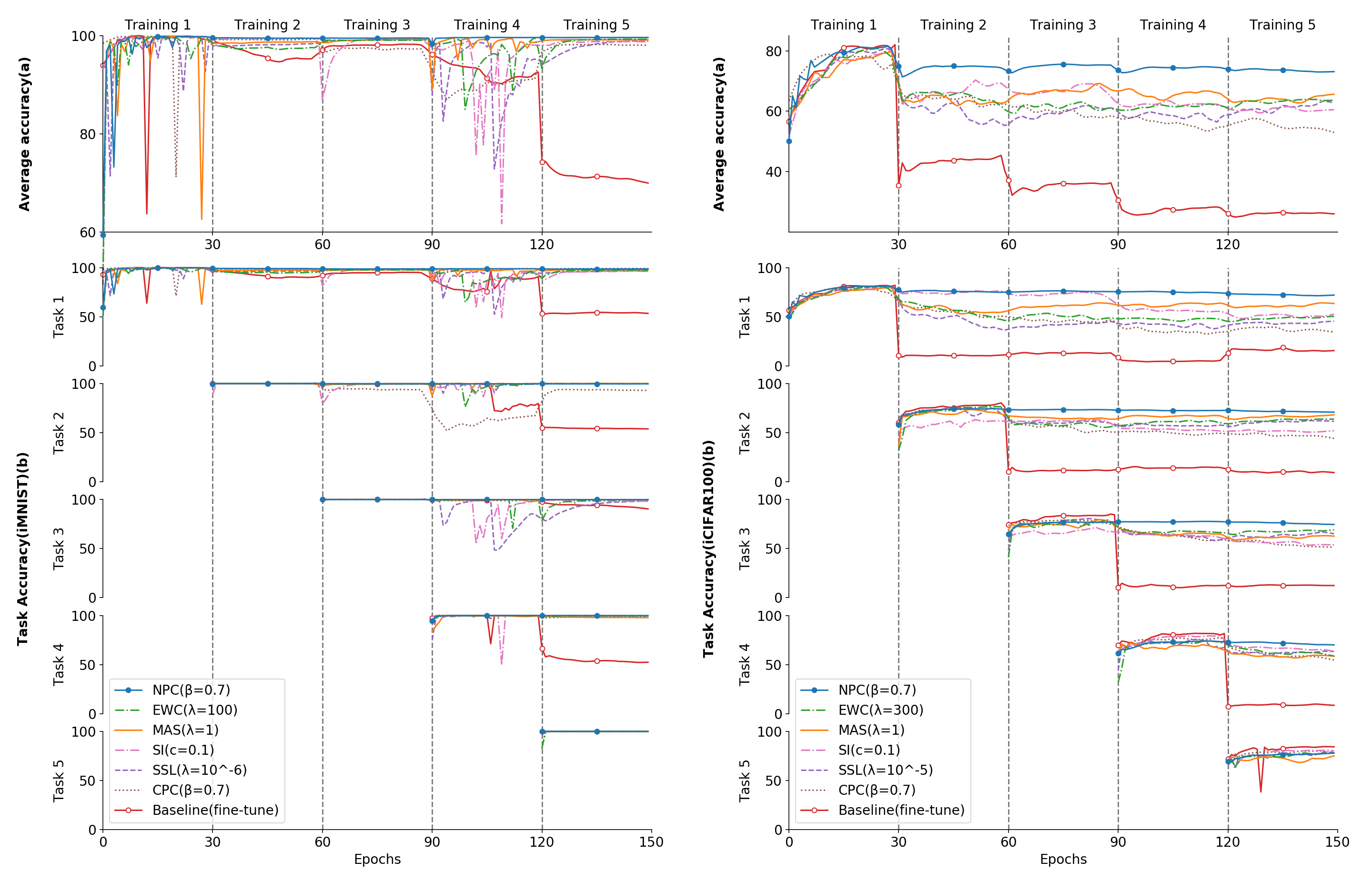}}
\caption{
    Training curves of continual learning algorithms on the iMNIST(left) and iCIFAR100(right) dataset. (a) Average validation accuracy of tasks that have been trained up to each moment. (b) Training curves of the five tasks according to the learning algorithms. See Table 1 for numerical results including average accuracy and standard error.
}
\label{training_curve}
\end{center}
\vskip -0.2in
\end{figure}

\begin{table}[H]
\begin{center}
\begin{tabular}{l|l|l|l|l|l|l}
\multirow{2}*{Alg.} & \multicolumn{6}{|c}{Validation accuracy on iMNIST dataset}\\ \cline{2-7}
& Task 1 & Task 2 & Task 3 & Task 4 & Task 5 & Average\\ \hline
NPC & \textbf{98.45(0.13)} & \textbf{99.87(0.12)} & \textbf{100.0(0.00)} & \textbf{99.85(0.06)} & \textbf{100.0(0.00)} & \textbf{99.63(0.02)}  \\ \hline
EWC & 87.60(8.70) & 99.14(0.49) & 99.52(0.22) & 90.84(8.69) & \textbf{100.0(0.00)} & 95.42(3.55) \\ \hline
MAS & 98.03(0.60) & 99.96(0.04) & 99.95(0.05) & 95.88(2.41) & \textbf{100.0(0.00)} & 98.76(0.38)  \\ \hline
SI & 94.65(0.88) & 99.48(0.14) & 92.41(6.23) & 99.67(0.12) & \textbf{100.0(0.00)} & 97.24(1.39) \\ \hline
SSL & 93.01(1.84) & 95.85(3.88) & 99.02(0.39) & 96.49(2.12) & \textbf{100.0(0.00)} & 96.87(1.08) \\ \hline
CPC & 96.95(0.69) & 92.70(0.39) & 99.07(0.43) & 93.72(0.59) & \textbf{100.0(0.00)} & 96.49(0.65)
\vspace*{0.5 cm}
\end{tabular}
\begin{tabular}{l|l|l|l|l|l|l}
\multirow{2}*{Alg.} & \multicolumn{6}{|c}{Validation accuracy on iCIFAR100 dataset}\\ \cline{2-7}
& Task 1 & Task 2 & Task 3 & Task 4 & Task 5 & Average\\ \hline
NPC & \textbf{69.20(1.35)} & \textbf{70.27(0.35)} & \textbf{74.33(0.15)} & \textbf{67.90(1.63)} & 77.67(0.24) & \textbf{71.87(0.65)} \\ \hline
EWC & 54.70(5.70) & 61.87(0.98) & 67.37(2.13) & 60.17(2.10) & 78.73(0.74) & 64.57(0.94) \\ \hline
MAS & 67.37(2.49) & 64.23(2.30) & 61.70(1.38) & 57.77(0.70) & 76.67(1.22) & 65.54(0.73)  \\ \hline
SI & 48.47(2.24) & 48.90(1.85) & 53.47(0.84) & 58.27(2.83) & \textbf{80.47(0.52)} & 57.91(1.16) \\ \hline
SSL & 58.47(6.20) & 66.27(2.32) & 61.90(1.86) & 61.43(0.62) & 77.70(1.23) & 65.16(1.37)\\ \hline
CPC & 25.97(4.13) & 42.33(1.87) & 52.60(1.40) & 53.97(0.64) & 78.60(1.10) & 50.70(1.27) \\
\end{tabular}
\end{center}
\caption{Average validation accuracies($\pm$standard error) of the 5 different continual learning algorithms on iMNIST and iCIFAR100 dataset after training all the 5 tasks. Note that we used different hyperparameters in each experiments in some cases to obtain the best performance of each algorithm. We used ($\beta_{NPC} =\beta_{CPC} = 0.7, \lambda_{EWC} = 100, \lambda_{MAS} = 1.0, \lambda_{SI} = 0.1, \lambda_{SSL} = 10^{-6}$) for iMNIST, and ($\lambda_{EWC} = 300, \lambda_{SSL} = 10^{-5}$) for iCIFAR100. NPC exhibits the best performance in all columns except for Task 5 of iCIFAR100 dataset.}
\end{table}

\begin{figure}[H]
\vskip 0.2in
\begin{center}
\centerline{\includegraphics[width=17cm]{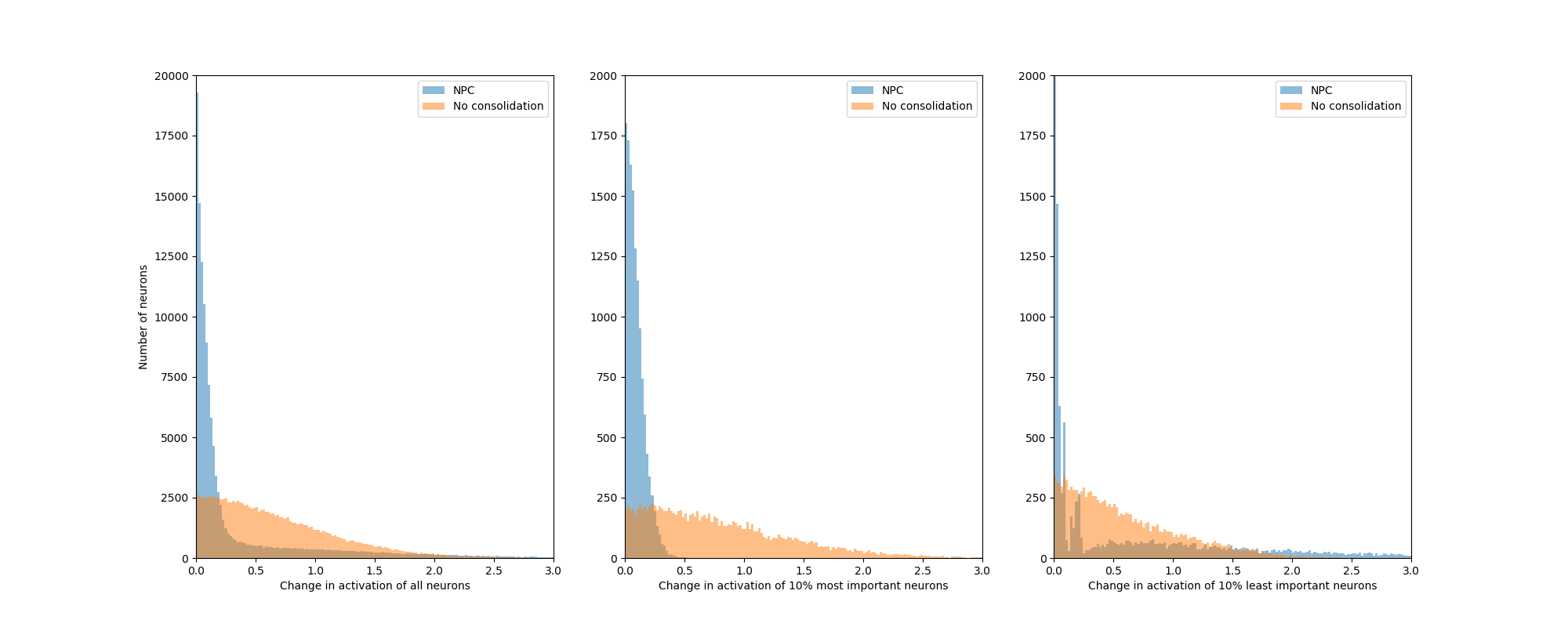}}
\caption{
    Change in the activation of the neurons on the second top layer after learning a subsequent task. Using NPC, the average change of the important neurons(0.094, center) was significantly smaller than the average change of all neurons(0.383, left), while the average change of less important neurons(0.667, right) was much larger than that of all neurons. Without consolidation, there was no meaningful correlation between the change in neuron activation and their importance values.
}
\label{histogram} 
\end{center}
\vskip -0.2in
\end{figure}

 Task 1, while Task 2 was learned mainly by the neurons less important for Task 1.

\section{Conclusion and Discussion}

In this paper, we proposed a continual learning algorithm, NPC, that controls the plasticity of a neural network at the neuron level, instead of restricting individual connection weights to stay close to certain values. NPC is effective in preserving old knowledge since it consolidates all the incoming pathways to important neurons. The experimental results on the iMNIST and iCIFAR100 datasets show that NPC is significantly more effective than conventional connection-level consolidation algorithms that do not consider the relation among connections. NPC has an additional benefit that it does not maintain any task-specific information such as the latest set of parameters optimized for each task, which makes it more efficient in terms of memory and computational complexity.


In applying NPC to more diverse architectures, the residual 
connection\citep{ResNet} is one of the hurdles that should be addressed. Interpreting the summation of multiple neuron outputs and determining which neurons should be preserved are non-trivial problems, especially when the important and unimportant neurons are added up by residual connection. 

While NPC defines the unit and the method for controlling plasticity, strategies for evaluating and managing the importance value of each neuron leaves room for exploration. Studies on network pruning show us how deep learning models can learn complicated knowledge with a surprisingly small size. However, without explicit intervention, deep neural networks tend to consume more capacity than actually needed. We believe that NPC will benefit greatly if there is a method to enforce the model to use minimal capacity per task.

\bibliographystyle{spbasic}

\end{document}